\documentclass[10pt,twocolumn,letterpaper]{article}

\usepackage[pagenumbers]{cvpr} %

\newcommand{\ours}{MAGiC-SLAM\xspace}
\newcommand{\boldparagraph}[1]{\vspace{2pt}\noindent{\bf #1}}

\usepackage{colortbl}
\usepackage{arydshln}
\usepackage{pifont}
\usepackage{xcolor}
\usepackage{multirow}
\usepackage{makecell}
\usepackage{tabularx}

\colorlet{colorFst}{Green!25}       %
\colorlet{colorSnd}{SpringGreen!45} %
\colorlet{colorTrd}{Yellow!30}      %
\colorlet{colorLow}{darkgray!30}    %
\newcommand{\redx}{{\color{red}\ding{55}}}
\newcommand{\fs}{\cellcolor{colorFst}\bf}   %
\newcommand{\nd}{\cellcolor{colorSnd}}      %
\newcommand{\rd}{\cellcolor{colorTrd}}      %

\definecolor{cvprblue}{rgb}{0.21,0.49,0.74}
\usepackage[pagebackref,breaklinks,colorlinks,allcolors=cvprblue]{hyperref}

\title{MAGiC-SLAM: Multi-Agent Gaussian Globally Consistent  SLAM}

\author{
    Vladimir Yugay \hspace{1cm} Theo Gevers \hspace{1cm} Martin R. Oswald\\
    University of Amsterdam, Netherlands\\
    {\tt\small \{vladimir.yugay, th.gevers, m.r.oswald\}@uva.nl}\\
    {\small \href{https://vladimiryugay.github.io/magic_slam}{\nolinkurl{vladimiryugay.github.io/magic_slam}}} 
}

\begin{document}
\twocolumn[{%
  \maketitle
  \captionsetup{type=figure} %
  \includegraphics[width=\textwidth]{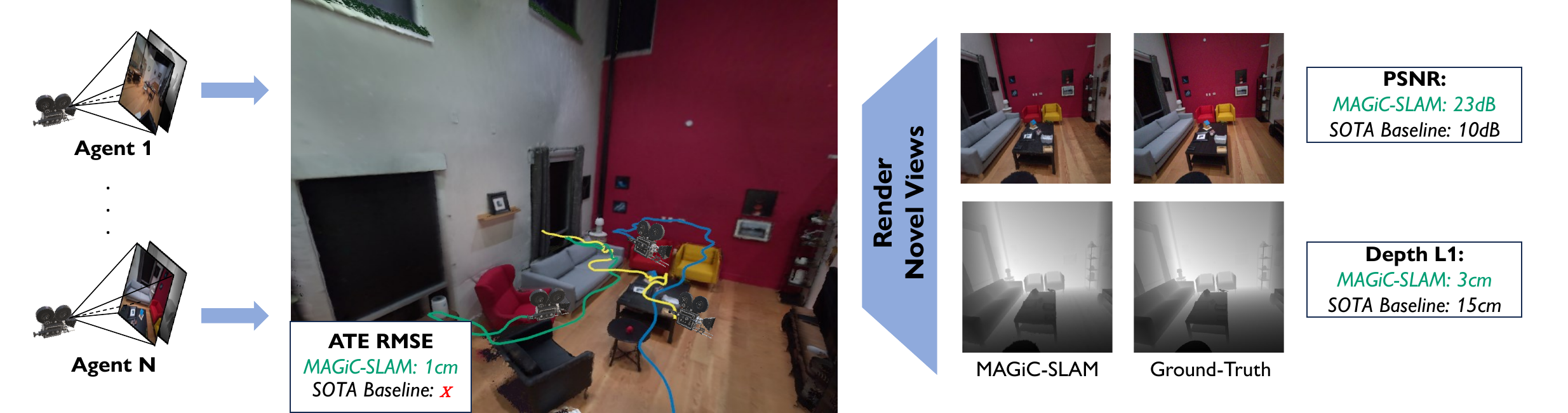}    
  \captionof{figure}{\textbf{\ours} is a multi-agent SLAM method capable of novel view synthesis. Given single-camera RGBD input streams from multiple simultaneously operating agents \ours estimates their trajectories and reconstructs a 3D Gaussian map that can be rendered from previously unseen viewpoints. We showcase the high-fidelity 3D Gaussian map of a real-world environment alongside multiple agent trajectories (depicted in green, yellow, and blue) within it. Our method effectively utilizes information from multiple agents to achieve centimeter-level tracking accuracy. Our mapping and map merging strategies allow for realistic rendering of color and depth, significantly improving the state of the art. Unlike previous methods, \ours is flexible in the number of agents it can handle.
  }
  \label{fig:teaser}
  \vspace{2em}
}]
\begin{abstract}
Simultaneous localization and mapping (SLAM) systems with novel view synthesis capabilities are widely used in computer vision, with applications in augmented reality, robotics, and autonomous driving. However, existing approaches are limited to single-agent operation. Recent work has addressed this problem using a distributed neural scene representation. Unfortunately, existing methods are slow, cannot accurately render real-world data, are restricted to two agents, and have limited tracking accuracy. In contrast, we propose a rigidly deformable 3D Gaussian-based scene representation that dramatically speeds up the system. However, improving tracking accuracy and reconstructing a globally consistent map from multiple agents remains challenging due to trajectory drift and discrepancies across agents' observations. Therefore, we propose new tracking and map-merging mechanisms and integrate loop closure in the Gaussian-based SLAM pipeline. We evaluate \ours on synthetic and real-world datasets and find it more accurate and faster than the state of the art.
\end{abstract}

\section{Introduction} 

Visual Simultaneous Localization and Mapping (SLAM) enables a machine to build a 3D map of its surroundings while determining its location relying solely on its camera input. Imagine an autonomous car navigating a busy city or a robot exploring a cluttered room. For these systems to move safely and make decisions, they need a reliable understanding of their surroundings. Over the years, visual SLAM has advanced~\cite{fuentes2015visual,kazerouni2022survey,tosi2024nerfs3dgaussiansplatting}, learning to handle complex scenes with increasing accuracy. This capability forms the backbone for various systems from self-driving cars to virtual reality glasses, where spatial understanding is crucial for navigation, planning, and interactive experiences.

Recently, SLAM systems have been enhanced with novel view synthesis (NVS) capabilities~\cite{Sucar2021IMAP:Real-Time,Matsuki:Murai:etal:CVPR2024,keetha2023splatam}, allowing them to generate realistic, immersive views of scenes. This allows more detailed scene exploration and supports the creation of high-quality virtual environments. However, these NVS-capable SLAM methods are
slow, and tracking accuracy remains limited~\cite{keetha2023splatam}. These limitations become even more pronounced in a multi-agent setup, with larger scenes and numerous observations.

Effectively processing information from multiple agents operating simultaneously opens up new possibilities for SLAM systems. First, this naturally accelerates 3D reconstruction by enabling horizontal parallelization; each agent can map different parts of the environment concurrently, leading to faster, more efficient coverage of large areas~\cite{chang21icra_kimeramulti}. Second, agents can collaboratively refine their location estimates by sharing their observations~\cite{lajoieSwarmSLAM}. Finally, the global map constructed from the perspectives of multiple agents is more geometrically consistent and accurate~\cite{schmuck2017ccm}. 

The only solution for multi-agent NVS-capable SLAM hitherto has combined a distributed neural scene representation with a traditional loop closure mechanism~\cite{hu2023cpslamcollaborativeneuralpointbased}. However, this approach faces several issues. Firstly, it fails to reconstruct a coherent map capable of accurate NVS. Neural scene representations inherently lack support for rigid body transformations~\cite{liso2024loopy} making it impossible to update and merge the global map efficiently. This in turn leads to poor rendering quality of the novel views. Secondly, localization accuracy remains limited, as neural-based camera pose optimization struggles with the variability and imperfections inherent to real-world data. Finally, the system is prohibitively slow, demands extensive computational resources, and supports only two agents.

We present \ours, a multi-agent NVS-capable SLAM pipeline designed to overcome these limitations. Firstly, \ours utilizes 3D Gaussians as the scene representation supporting rigid body transformations. Every agent processes the input RGBD sequence in chunks (sub-maps) which can be effectively corrected and merged into a coherent global map. Secondly, we implement a loop closure mechanism to improve trajectory accuracy by leveraging information from all agents. It is further enhanced with a novel loop detection module, based on a foundational vision model, which enables better generalization to unseen environments. Finally, our flexible tracking and mapping modules allow our system to achieve superior speed and easily scale to varying numbers of agents.

Overall, our contributions can be summarized as follows:
\begin{itemize}
    \item A multi-agent NVS-capable SLAM system for consistent 3D reconstruction supporting an arbitrary number of simultaneously operating agents
    \item A loop closure mechanism for Gaussian maps leveraging a foundational vision model for loop detection
    \item Efficient map optimization and fusion strategies that reduce required disk storage and processing time
    \item A robust Gaussian-based tracking module
\end{itemize}

\section{Related Work} \label{section:related_work}

\boldparagraph{Neural SLAM.}
Neural radiance fields (NeRF)~\cite{Mildenhall2020NeRF:Synthesis} have achieved remarkable results
in novel view synthesis~\cite{mipnerf,plenoxels,muller2022instant}.
\cite{Sucar2021IMAP:Real-Time} started a whole new line of research~\cite{zhu2022nice,yang2022vox,wang2023co,mahdi2022eslam,sandstrom2023point,
liso2024loopy,zhang2023go} by proposing a SLAM system using neural fields to represent the map and optimize the camera poses. Various neural
 field approaches have provided insights for neural SLAM developments. For instance, NICE-SLAM~\cite{zhu2022nice} uses a voxel grid to store neural features,
while Vox-Fusion~\cite{yang2022vox} improves the grid with adaptive sizing. Point-SLAM~\cite{sandstrom2023point} attaches feature embeddings to point
clouds on object surfaces, offering more flexibility and the ability to encode concentrated volume density.
Co-SLAM~\cite{wang2023co} employs a hybrid representation combining coordinate encoding and hash grids to achieve smoother
reconstruction and faster convergence. A group of methods~\cite{Rosinol2022NeRF-SLAM:Fields,chung2022orbeez} use neural fields solely for
mapping while relying on traditional feature point-based visual odometry for tracking. Loopy-SLAM~\cite{liso2024loopy} explores the usage
of pose-graph optimization to handle trajectory drift. While these methods have shown success in NVS, they are computationally intensive, slow, and struggle to render real-world environments accurately~\cite{keetha2023splatam,yugay2024gaussianslamphotorealisticdenseslam}. Additionally, neural maps inherently lack support for rigid body transformations~\cite{liso2024loopy}, which greatly limits the efficiency of map correction. In contrast, our scene representation is fast to render and optimize, significantly accelerating our SLAM pipeline. With native support for rigid body transformations, our approach enables faster map updates and seamless map merging.

\boldparagraph{Gaussian SLAM.}
Recently, 3D Gaussian Splatting(3DGS)~\cite{kerbl3Dgaussians} revolutionized novel view synthesis, achieving photorealistic real-time rendering at over 100 FPS without relying on neural networks. Compared to NeRFs, 3DGS is more efficient in terms of memory, and faster to optimize.
These factors inspired the line of SLAM
systems~\cite{Matsuki:Murai:etal:CVPR2024,keetha2023splatam,yan2024gsslamdensevisualslam,hhuang2024photoslam,yugay2024gaussianslamphotorealisticdenseslam,
hu2024cgslamefficientdensergbd} using 3D Gaussians instead of neural fields for tracking and mapping. \cite{Matsuki:Murai:etal:CVPR2024,yan2024gsslamdensevisualslam,hu2024cgslamefficientdensergbd} estimated camera pose by computing camera gradients from the 3DGS field analytically.
Others~\cite{keetha2023splatam,yugay2024gaussianslamphotorealisticdenseslam} design a warp-based camera pose optimization.
Finally,~\cite{peng2024rtgslam,hhuang2024photoslam} use existing sparse SLAM systems to speed up tracking. While addressing many limitations of neural SLAM, the proposed methods overlook global consistency, leading to trajectory error accumulation~\cite{loop_closure}. In contrast, \ours integrates a loop closure mechanism within the 3DGS SLAM pipeline to correct accumulated trajectory errors. It effectively generalizes to unknown environments leveraging a foundational vision model for loop detection.
Concurrently, Zhu et al.~\cite{zhu2024loopsplatloopclosureregistering} proposed a way to do loop closure in a 3DGS-based SLAM pipeline. However, the method cannot handle multiple agents, uses a standard loop closure detection mechanism, and is less efficient in tracking and mapping.

\boldparagraph{Multi-agent Visual SLAM.}
Despite significant progress in single-agent systems, multi-agent SLAM remains less developed due to the complexity of designing a multi-agent pipeline. Multi-agent SLAM can be categorized into two types: centralized and distributed. In a distributed system~\cite{Lajoie_2020,chang21icra_kimeramulti} agents communicate sporadically. While being more robust in environments with stringent networking constraints, their performance is limited by the computing power of a separate agent. Centralized systems~\cite{schmuck2017ccm, lajoieSwarmSLAM} use a centralized server to manage agents' maps and globally optimize their trajectories. Despite the success of these methods, they do not provide a map with novel view synthesis capabilities. Hu et al.~\cite{hu2023cpslamcollaborativeneuralpointbased} proposed a centralized neural-based multi-agent NVS-capable SLAM to address this. However, it inherited all the problems associated with neural representations such as low speed, compute requirements, and ability to render real-world data~\cite{keetha2023splatam}. In addition, CP-SLAM~\cite{hu2023cpslamcollaborativeneuralpointbased}, can only support two agents operating simultaneously. In contrast, \ours employs an efficient scene representation that enables seamless global map reconstruction with NVS capabilities suited for real-world environments. Additionally, it delivers improved tracking accuracy through robust tracking and loop closure modules. Finally, our system is flexible in handling multiple agents, constrained only by the capacity of the centralized server.

\section{Method} \label{method}

\begin{figure*}[ht!]
  \centering
  \includegraphics[width=\linewidth]{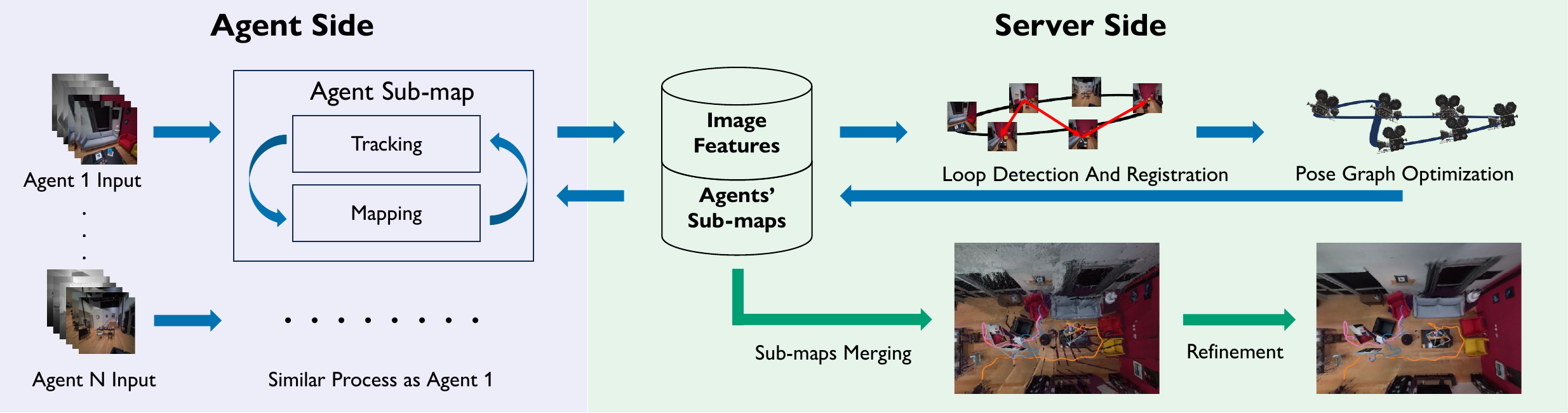}\\
  \caption{\textbf{\ours{} Architecture.} \textit{Agent Side:} Each agent processes a separate RGBD stream, maintaining a local sub-map and estimating its trajectory. When an agent starts a new sub-map, it sends the previous sub-map and image features to the centralized server. \textit{Server Side:} The server stores the image features and sub-maps from all agents and performs loop closure detection, loop constraint estimation, and pose graph optimization. It then updates the stored sub-maps and returns the optimized poses to the agents. Once the algorithm completes (denoted by green arrows), the server merges the accumulated sub-maps into a single unified map and refines it.}
  \label{fig:architecture}
\end{figure*}

We introduce \ours, which architecture is shown in Figure~\ref {fig:architecture}. Every agent processes an RGB-D stream performing local mapping and tracking. 3D Gaussians are used to represent the agents' sub-maps and to improve tracking accuracy. The foundation vision model-based loop detection module extracts features from RGB images and sends them to the centralized server with the local sub-map data.
The server detects the loops based on the image encodings, performs pose graph optimization, and sends the optimized poses back to the agents.
At the end of the run, the server fuses the agents' sub-maps into a global Gaussian map. This section introduces per-agent mapping and tracking mechanisms and describes global loop closure and map construction processes.

\subsection{Mapping} \label{method:mapping}

Every agent processes a single sub-map of a limited size, represented as a collection of 3D Gaussians.
Sub-maps are initialized from the first frame. $\theta_\text{sample}$ points sampled from the lifted to 3D RGBD frame serve as the means for new 3D Gaussians. New Gaussians are added to regions of the active sub-map with low Gaussian density, based on rendered opacity, and are optimized using the loss:
\begin{equation}
    L_\text{mapping} = \lambda_\text{color} \cdot L_\text{color} + \lambda_\text{depth} \cdot L_\text{depth} + \lambda_\text{reg} \cdot L_\text{reg} \enspace,
\label{eq:joint_loss}
\end{equation}
where are $\lambda_{*} \in R$ are hyperparameters. The color loss $L_\text{color}$ is is defined as:
\begin{equation}
    L_\text{color} = (1 - \lambda) \cdot |\hat{I} - I|_1 + 
       \lambda \big(1 - \mathrm{SSIM}(\hat{I}, I) \big) \enspace,
    \label{eq:color_loss}
\end{equation}
where $\hat{I}, I$ are the rendered and input images respectively, and  $\lambda \in R$ is the weighting factor.
Depth loss $L_\text{depth}$ is formulated as:
\begin{equation}
    L_\text{depth} = |\hat{D} - D|_1 \enspace,
    \label{eq:depth_loss}
\end{equation}
where $\hat{D}, D$ are the rendered and input depth maps.
The regularization loss $L_\text{reg}$ is represented as:
\begin{equation}
    L_\text{reg} = |K|^{-1} \sum_{k \in K}|s_k - \overline{s}_k|_1,
\end{equation}
where $s_k\in\mathbb{R}^3$ is the scale of a 3D Gaussian, $\overline{s}_k$ is the mean sub-map scale,
and $|K|$ is the number of Gaussians in the sub-map. We do not optimize the spherical harmonics of the Gaussians to reduce their memory footprint and improve tracking accuracy~\cite{Matsuki:Murai:etal:CVPR2024}.

A new sub-map is created, and the previous one is sent to the server after every $\theta_\text{submap}$ frame. Creating new sub-maps in already mapped areas introduces some computational redundancy, but it limits overall computational cost and maintains tracking and mapping speed as the scene expands.
Moreover, unlike in ~\cite{yugay2024gaussianslamphotorealisticdenseslam}, only Gaussians with zero rendered opacity in the current camera frustum are dispatched to the server.
This approach significantly reduces the disk space required to store the sub-maps and speeds up the map merging process. The supplementary material provides a quantitative evaluation of the strategy.

\subsection{Tracking} \label{method:tracking}

Typically GS SLAM systems optimize camera pose either explicitly or implicitly. In the explicit case, the camera pose gradient
is derived~\cite{matsuki2023newton} or approximated~\cite{yan2024gsslamdensevisualslam, hu2024cgslamefficientdensergbd} analytically. In the implicit case~\cite{keetha2023splatam,yugay2024gaussianslamphotorealisticdenseslam}, the relative pose between two frames is optimized by warping the GS point clouds and minimizing re-rendering depth and color errors. In both cases, the camera pose is estimated based on the existing map, following a frame-to-model paradigm~\cite{kazerouni2022survey}.

Considering the advantages and disadvantages of frame-to-frame and frame-to-model tracking paradigms~\cite{fuentes2015visual,kazerouni2022survey,tosi2024nerfs3dgaussiansplatting} and the convenience of using sub-maps for loop closure, we propose a hybrid implicit tracking approach that combines the strengths of both. Specifically, we initialize the relative pose using a deterministic frame-to-frame dense registration and then refine it through frame-to-model optimization. This approach differs from previous NVS SLAM methods, initializing the relative pose based on a constant speed assumption.

Moreover, we found implicit tracking to be more accurate than the explicit approach given robust pose initialization. At the same time, explicit tracking methods do not benefit from pose initialization since camera poses are optimized across the whole co-visibility window at every mapping step. We refer to supplementary material for experiments highlighting this phenomenon.

\boldparagraph{Pose Initialization.}
At time $t$, given the input colored point cloud $P_t$, the goal is to estimate its pose $T_{t-1, t} \in SE(3)$ relative to the previous input point cloud $P_{t - 1}$.  The registration process is performed iteratively at several scales, 
starting from the coarsest, $l = 0$, down to the finest, $l = L$. At every scale, both point clouds are voxelized, and the set of correspondences $K = \{(p, q)\}$ between $P_{t}, P_{t-1}$ is computed using ICP~\cite{icp}. For every scale, a loss function:
\begin{align}
    E(\mathbf{T_{t-1, t}}) = (1 - \sigma) \sum_{(\mathbf{p}, \mathbf{q}) \in \mathcal{K}} \left( r^{(\mathbf{p}, \mathbf{q})}_C(\mathbf{T_{t-1,t}}) \right)^2 \\\nonumber
    + \sigma \sum_{(\mathbf{p}, \mathbf{q}) \in \mathcal{K}} \left( r^{(\mathbf{p}, \mathbf{q})}_G(\mathbf{T_{t-1,t}}) \right)^2,
\end{align}
is optimized where $\sigma \in [0, 1]$ is a scalar weight. $r_G^{(p,q)}$ is a geometric residual defined as:
\begin{equation}
r_G^{(p,q)} = \Big(\big(T_{t-1, t}(q) - p\big)n_{p}\Big)^{2}.
\end{equation}
where $n_p$ is a normal vector of $p$.
The color residual $r_C^{(p,q)}$ is computed as:
\begin{equation}
r_C^{(p,q)} = C_p\Big(f_{p}\big(T_{t,t-1}(q)\big) - p\Big) - C(q),
\end{equation}
where $f_{p}$ is a function projecting a point on the tangent plane of $p$, $C(q)$ is the intensity of point $q$, and $C_{p}$ is a continuous intensity function defined over point cloud $P_{t}$. The loss is optimized until convergence with the Gauss-Newton method. For more details about the registration mechanism please refer to~\cite{8237287}.

\boldparagraph{Pose Refinement.} The pose obtained in the initialization step is further refined using re-rendering losses~\cite{yugay2024gaussianslamphotorealisticdenseslam} of the scene. To refine the initial pose estimate, we freeze all Gaussian parameters and minimize the loss:
\begin{equation}
    \underset{T_{t - 1, t}}{\mathrm{arg\,min}} \, L_\text{tracking}\Big(\hat{I}(T_{t - 1, t}), \hat{D}(T_{t - 1, t}), I_{t}, D_{t}, \alpha_t \Big) \enspace,
\end{equation}
where $\hat{I}(T_{t - 1, t})$ and $\hat{D}(T_{t - 1, t})$ are the rendered color and depth from the sub-map warped with the relative transformation
$T_{t-1, t}$, $I_{t}$ and $D_{t}$ are the input color and depth map at frame $t$.

We use soft alpha and color rendering error masking to avoid contaminating the tracking loss with pixels from previously unobserved or poorly reconstructed areas~\cite{keetha2023splatam,yugay2024gaussianslamphotorealisticdenseslam}.
Soft alpha mask $M_\text{alpha}$ is a polynomial of the alpha map rendered from the 3D Gaussians.
Error boolean mask $M_\text{inlier}$ discards all the pixels where the color and depth errors are larger than a frame-relative error threshold:
\begin{equation}
    L_\text{tracking} = \!\sum\! M_\text{inlier} \cdot M_\text{alpha} \cdot\big(\lambda_{c}|\hat{I} - I|_1 + (1 - \lambda_{c})|\hat{D} - D|_1\big).
\end{equation}
The weighting ensures the optimization is guided by well-reconstructed regions where the accumulated alpha
values are close to 1 and rendering quality is high.

\subsection{Loop Closure} \label{method:loop_closure}

Loop closure is the process that detects when a system revisits a previously mapped area and adjusts the map and camera poses to reduce accumulated drift, ensuring global consistency. It includes four key steps: loop detection, loop constraint estimation, pose graph optimization, and integration of optimized poses. Loop detection identifies when a previously mapped location is revisited. Loop constraint estimation calculates the relative pose between frames in the loop. Pose graph optimization then refines all camera poses, minimizing discrepancies between odometry and loop constraints to maintain global consistency. Finally, the optimized poses are integrated into the reconstructed map.

\boldparagraph{Loop Detection.} Throughout the run, each agent extracts features from the first frame in each sub-map and sends them to a centralized server. The features are stored in a GPU database~\cite{douze2024faisslibrary} optimized for similarity search. The server queries the database for potential loop candidates for every new sub-map frame. A pair of frames is considered a loop if the distance between them is less than a threshold $\theta_\text{feature}$ in the image feature space. Loops belonging to the same agent are additionally filtered based on time threshold $\theta_\text{time}$ to avoid too many uninformative loops.

Loop detection heavily influences the accuracy of loop edge constraint estimation since large frame overlap is crucial for registration~\cite{liso2024loopy}. Common approaches for loop closure detection are based on ORB~\cite{GalvezTRO12}
or NetVLAD~\cite{netvlad} image descriptors. Hu et al.~\cite{hu2023cpslamcollaborativeneuralpointbased} argue that NetVLAD global image descriptors are better suited for loop closure detection. However, NetVLAD is trained on a relatively small dataset which leads to a lack of generalization to unknown environments.
To overcome this limitation we propose to use a foundational vision model as a feature extractor. We use a small variation of DinoV2~\cite{oquab2024dinov2learningrobustvisual} because of the large amount of data it was trained on, its compactness, and the quality of
the features it produces for the downstream tasks.

\boldparagraph{Loop Constraints.} We use a coarse-to-fine registration approach to estimate loop edge constraints. For the coarse alignment, we apply the global registration method of Rusu et al.~\cite{rusu}, which extracts Fast Point Feature Histograms (FPFH) from downsampled versions of the source ($P_s$) and target ($P_t$) point clouds. Correspondence search is then performed in the FPFH feature space rather than in Euclidean space. Optimization is embedded in a RANSAC framework to reject outlier correspondences, producing a rigid transformation of the source point cloud $S_s$ to align with the target $S_t$. Finally, ICP~\cite{icp} is applied to the full-resolution point clouds to refine the coarse alignment estimate.

We found that directly registering Gaussian means from different agents is unreliable, as Gaussians representing overlapping regions can have widely varying distributions across agents. To address this, we anchor an input point cloud at the start of each sub-map and use it for registration.

\boldparagraph{Pose Graph Optimization.} Each node in a pose graph, $T_i \in SE(3)$, represents a distinct sub-map. Neighboring sub-maps are connected by odometry edges, derived from the tracker, representing the relative transformations between them. Loop edge constraints $T_{st} \in SE(3)$ are added between non-adjacent sub-maps and computed using a registration method different from that of the tracker. The error term between two nodes $i$ and $j$ is defined as:
\begin{equation}
    e_{ij} = \log\big(T_{ij}^{-1} (T_i^{-1} T_j)\big),
\end{equation}
where $T_{ij}$ is the relative transformation between the nodes $i$ and $j$, $T_i$, $T_j$ are the node poses, and
$\log$ is the logarithmic map from $SE(3)$ to $se(3)$. To get the optimized camera poses we minimize the error term:
\begin{equation}
    F(T) = \sum_{\langle i,j \rangle \in \mathcal{C}} 
    \big( e(T_i, T_j)^\top 
    \Omega_{ij} 
    e(T_i, T_j) \big)
\end{equation}
where $\Omega_{ij} \in R^{6 \times 6}$ is a positive semi-definite information matrix reflecting the uncertainty
of the constraint estimate - the higher the confidence of an edge, the larger the weight is applied to the residual.
The error terms are linearized using a first-order Taylor expansion, and the loss is minimized with the Gauss-Newton method. For further details, please see \cite{g2o}.

\boldparagraph{Pose Update Integration.} The pose graph optimization module provides pose corrections $\{T^i_c \in SE(3)\}_{i=1}^{N_{s}}$ for every sub-map of every agent. All camera poses $\{T^i_j\}_{j=1}^{N_{p}}$ belonging to a sub-map $i$ are corrected as:
\begin{equation}
    T^i_j \leftarrow T^c_i T^i_j.
\end{equation}
All Gaussians $\{G^i_j(\mu^i_j, \Sigma^i_j, o^i_j, c^i_j)\}_{j=1}^{N_{g}}$ belonging to sub-map $i$ are updated as well:
\begin{equation}
    \mu^i_j \leftarrow T^c_i \mu^i_j, \quad \Sigma^i_j \leftarrow T^c_{i, R} \Sigma^i_j,
\end{equation}
where $T^c_{i, R}$ is a rotation component of $T^c_i \in SE(3)$. We do not correct Gaussian colors since we do not optimize spherical harmonics (Subsection~\ref{method:mapping}).

\subsection{Global Map Construction} \label{method:fusion} 
\begin{figure}
\newcommand{\sz}{0.235}
\begin{subfigure}{\sz\textwidth}
  \centering
  \includegraphics[width=.99\linewidth]{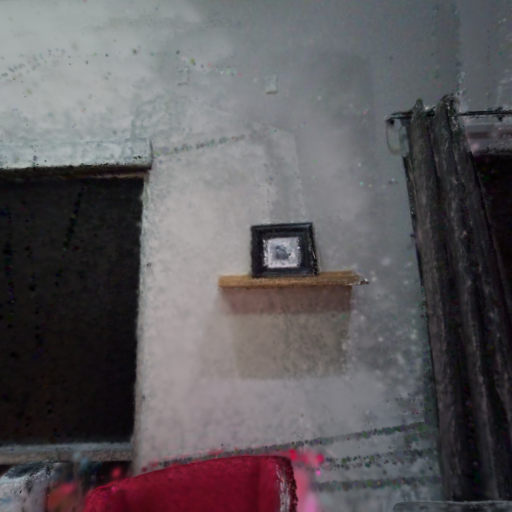}
  \caption{Visual artifacts}
  \label{fig:submap_artifacts}
\end{subfigure}
\begin{subfigure}{\sz\textwidth}
  \centering
  \includegraphics[width=.99\linewidth]{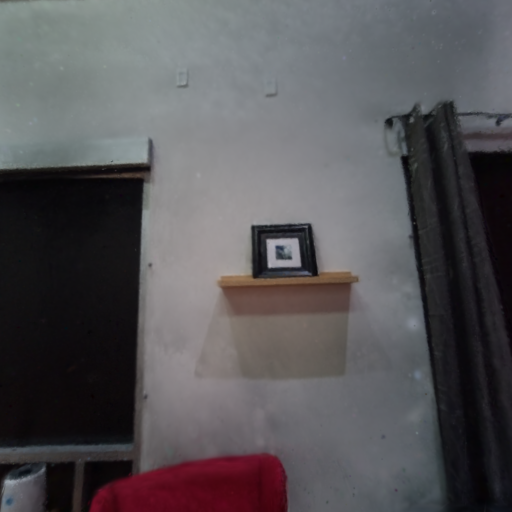}
  \caption{Refined view}
  \label{fig:refined_view_1}
\end{subfigure}

\begin{subfigure}{\sz\textwidth}
  \centering
  \includegraphics[width=.99\linewidth]{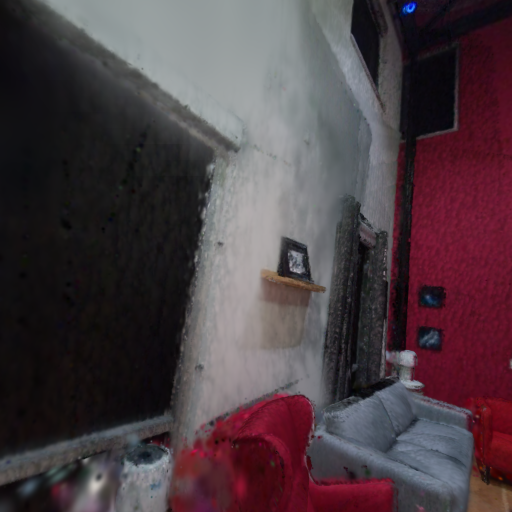}
  \caption{Geometric artifacts}
  \label{fig:geometric_artifacts}
\end{subfigure}
\begin{subfigure}{\sz\textwidth}
  \centering
  \includegraphics[width=.99\linewidth]{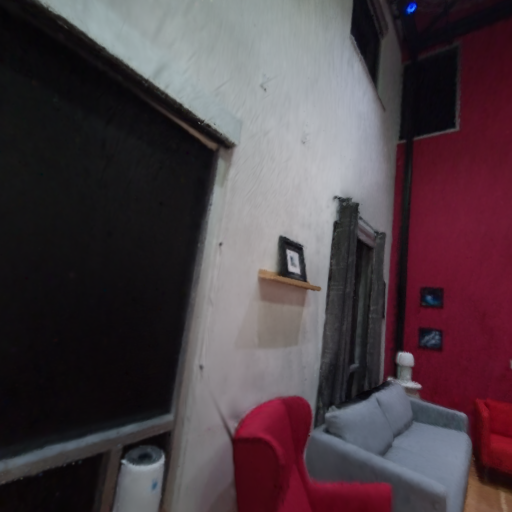}
  \caption{Refined view}
  \label{fig:refined_view_2}
\end{subfigure}

\caption{\textbf{Map Merging.}  Our coarse-to-fine strategy effectively removes (a) visual artifacts caused by the GS mechanism and (c) geometric artifacts resulting from Gaussian sub-map intersections.}

\label{fig:map_merging}
\vspace{-10pt}
\end{figure}

Once the agents complete processing their data, the server merges the sub-maps from all agents into a unified global map. The map is merged in two stages: coarse and fine. During the coarse stage, the server loads the cached sub-maps and appends them into a single global map. Caching Gaussians that are not visible from the first keyframe of the next sub-map allows for appending the Gaussians without the need for costly intersection check~\cite{yugay2024gaussianslamphotorealisticdenseslam}. However, this results in visual artifacts at the edges of the renderings. This happens because some Gaussians with zero opacity for a given view still influence the rendering through the projected 2D densities at the edges.
Additionally, the coarse merging of sub-maps might introduce geometric artifacts at their intersections. The fine merging stage addresses these issues by optimizing the Gaussian parameters using color and depth rendering losses for a small number of iterations and pruning Gaussians with zero opacity. The visual effects of the refining step are shown in Fig.~\ref{fig:map_merging}.

\section{Experiments}
We describe our experimental setup and compare \ours with state-of-the-art baselines. We assess tracking and rendering performance on synthetic and real-world multi-agent datasets and provide ablation studies on key components of our pipeline. Please refer to the supplementary material for the implementation details and hyperparameters for tracking, mapping, and loop closure modules.

\boldparagraph{Baselines.} To evaluate tracking, following~\cite{hu2023cpslamcollaborativeneuralpointbased} we compare our method to the state-of-the-art multi-agent
SLAM systems like SWARM-SLAM~\cite{lajoieSwarmSLAM}, CCM-SLAM~\cite{schmuck2017ccm}, and CP-SLAM~\cite{hu2023cpslamcollaborativeneuralpointbased}.
To make the comparison more comprehensive, we include several single-agent systems like Gaussian-SLAM~\cite{yugay2024gaussianslamphotorealisticdenseslam},
MonoGS~\cite{Matsuki:Murai:etal:CVPR2024}, and Orb-SLAM3~\cite{Campos_2021}. We include Orb-SLAM3 since it is one of the most popular
and reliable single-agent SLAM methods, Gaussian-SLAM since our approach uses sub-maps, and MonoGS as the most accurate
single-agent NVS SLAM system~\cite{tosi2024nerfs3dgaussiansplatting}. We evaluate rendering against CP-SLAM~\cite{hu2023cpslamcollaborativeneuralpointbased} since it is the only existing multi-agent NVS-capable SLAM system.

\boldparagraph{Datasets.} We test our method on the MultiagentReplica~\cite{hu2023cpslamcollaborativeneuralpointbased} dataset which contains four two-agent RGB-D sequences in a synthetic environment. Each sequence consists of 2500 frames, except for the Office-0 scene, which has 1500 frames.
MultiagentReplica is a synthetic dataset and does not have a test set for novel view synthesis evaluation. Therefore, we extend our evaluation to real-world scenes using the ego-centric Aria~\cite{pan2023ariadigitaltwinnew} dataset. The Aria dataset provides ground-truth depth and camera information from recordings of two rooms within a real-world environment. We selected this dataset for its high-quality ground-truth data and the growing prevalence of egocentric wearable devices. However, the original dataset includes many dynamic objects, and since our method is not designed for dynamic environments, this restricted the amount of usable data. We selected three sequences from every room to simulate multi-agent operations, choosing sequences with sufficient consecutive frames without dynamic objects. We then extracted 500 consecutive frames from each sequence for training. We sampled 100 unseen frames from each room to test our method's NVS capabilities. We refer to this dataset as AriaMultiagent in all tables and figures.

\boldparagraph{Evaluation Metrics.} To assess tracking accuracy, we use ATE RMSE~\citep{Sturm2012ASystems}, and for rendering we compute PSNR,
SSIM~\citep{wang2004image} and LPIPS~\citep{zhang2018unreasonable}. Rendering metrics on ReplicaMultiagent are evaluated by rendering full-resolution
images along the estimated trajectory with mapping intervals similar to~\cite {sandstrom2023point}. The same metrics over training and hold-out test frames
are used for the AriaMultiagent dataset. We evaluate the depth error using the L1 norm in centimeters.
\subsection{Tracking Performance}
In Table~\ref{tab:tracking_replica_multiagent} we compare our method against state-of-the-art multi-agent SLAM systems on
the ReplicaMultiagent dataset processing two agents simultaneously. In Table~\ref{tab:tracking_aria_multiagent} we evaluate
\ours on the AriaMultiagent dataset processing three agents simultaneously. Since CP-SLAM does not support the operation
of more than two agents, we report its results on the first two out of three agents.
To make our evaluation more comprehensive, we compare our method against single-agent SLAM systems in Table~\ref{tab:tracking_vs_single}.
We run single-agent systems on each agent separately and report their average performance. \ours achieves superior tracking accuracy thanks to our novel two-stage tracking mechanism. The loop closure mechanism effectively utilizes multi-agent information to enhance accuracy even further.

\begin{table}[ht]
    \centering
    \resizebox{\columnwidth}{!}{%
    \begin{tabular}{@{}llcccc@{}}
    \toprule
    \textbf{Method} & \textbf{Agent} & \textbf{Off-0} & \textbf{Apt-0} & \textbf{Apt-1} & \textbf{Apt-2} \\
    \midrule
    CCM-SLAM~\cite{schmuck2017ccm} &       \textbf{Agent 1}   & 9.84 & \redx & 2.12 & 0.51 \\
    Swarm-SLAM~\cite{lajoieSwarmSLAM} &               & 1.07 & 1.61 & 4.62 & 2.69 \\
    CP-SLAM~\cite{hu2023cpslamcollaborativeneuralpointbased} &                  & 0.50 & 0.62 & 1.11 & 1.41 \\
    \ours (w.o. Loop Closure) &         & 0.44 & 0.30 & 0.48 & 0.91 \\
    \ours &                    & 0.31 & 0.13 & 0.21 & 0.42 \\    
    \midrule
    CCM-SLAM~\cite{schmuck2017ccm} &        \textbf{Agent 2}  & 0.76 & \redx & 9.31 & 0.48 \\
    Swarm-SLAM~\cite{lajoieSwarmSLAM} &        & 1.76 & 1.98 & 6.50 & 8.53 \\
    CP-SLAM~\cite{hu2023cpslamcollaborativeneuralpointbased} &           & 0.79 & 1.28 & 1.72 & 2.41 \\
    \ours (w.o. Loop Closure) &  & 0.41 & 0.46 & 0.61 & 0.41 \\
    \ours &             & 0.24 & 0.21 & 0.30 & 0.22 \\    
    \midrule
    CCM-SLAM~\cite{schmuck2017ccm} &      \textbf{Average} & 5.30 & \redx & 5.71 & \nd0.49 \\
    Swarm-SLAM~\cite{lajoieSwarmSLAM} &     & 1.42 & 1.80 & 5.56 & 5.61 \\
    CP-SLAM~\cite{hu2023cpslamcollaborativeneuralpointbased} &        & \rd0.65 & \rd0.95 & \rd1.42 & 1.91 \\
    \ours (w.o. Loop Closure) &  & \nd0.42 & \nd0.38 & \nd0.54 & \rd0.66 \\
    \ours &          & \fs0.27 & \fs0.16 & \fs0.26 & \fs0.32 \\
    \bottomrule
    \end{tabular}%
    }
    \caption{\textbf{Tracking performance on ReplicaMultiagent~\cite{hu2023cpslamcollaborativeneuralpointbased} dataset}
    (ATE RMSE [cm]$\downarrow$). Comparison between \ours(w.o. Loop Closure) and \ours reveals the importance of
    loop closure for trajectory consistency. \redx{} indicates invalid results due to the failure of CCM-SLAM.}
    \label{tab:tracking_replica_multiagent}
\end{table}

\begin{table}[ht]
    \centering
    \resizebox{\columnwidth}{!}{%
    \begin{tabular}{@{}llcc@{}}
    \toprule
    \textbf{Method} & \textbf{Agent} & \textbf{Room0} & \textbf{Room1} \\
    \midrule
    CCM-SLAM~\cite{schmuck2017ccm} &    \textbf{Agent 1}      & \redx & \redx \\
    Swarm-SLAM~\cite{lajoieSwarmSLAM} &       & 6.11 & 4.29 \\
    CP-SLAM~\cite{hu2023cpslamcollaborativeneuralpointbased} &          & 0.68 & 5.06 \\
    \ours (w.o. Loop Closure) & & 0.92 & 1.15 \\
    \ours &   & 0.67 & 0.96 \\      
    \midrule
    CCM-SLAM~\cite{schmuck2017ccm} &   \textbf{Agent 2}         & \redx & \redx \\
    Swarm-SLAM~\cite{lajoieSwarmSLAM} &         & 8.43 & 4.95 \\
    CP-SLAM~\cite{hu2023cpslamcollaborativeneuralpointbased} &            & 5.39 & 0.68 \\
    \ours (w.o. Loop Closure) &   & 1.72 & 1.33 \\
    \ours &     & 1.13 & 0.53 \\    
    \midrule
    CCM-SLAM~\cite{schmuck2017ccm} &   \textbf{Agent 3}        & \redx & \redx \\
    Swarm-SLAM~\cite{lajoieSwarmSLAM} &        & 4.82 & 5.12 \\
    CP-SLAM~\cite{hu2023cpslamcollaborativeneuralpointbased} &           & \textbf{--} & \textbf{--} \\
    \ours (w.o. Loop Closure) &  & 3.76 & 1.25 \\
    \ours &    & 1.67 & 0.46 \\
    \midrule
    CCM-SLAM~\cite{schmuck2017ccm} &  \textbf{Average}                & \redx & \redx \\
    Swarm-SLAM~\cite{lajoieSwarmSLAM} &               & 6.45 & 4.78 \\
    CP-SLAM~\cite{hu2023cpslamcollaborativeneuralpointbased} &                  & \rd3.03 & \rd2.87 \\
    \ours (w.o. Loop Closure) &  & \nd2.13 & \nd1.24 \\
    \ours &           & \fs1.15 & \fs0.65 \\
    \bottomrule
    \end{tabular}%
    }
    \caption{\textbf{Tracking performance on AriaMultiagent dataset}
    (ATE RMSE [cm]$\downarrow$). Our method shows strong performance on real-world data. \redx{} indicates invalid results due to the
    failure of CCM-SLAM. \textbf{--} indicates that CP-SLAM does not support the operation of more than two agents.}
    \label{tab:tracking_aria_multiagent}
\end{table}

\begin{table}[ht]
    \centering
    \setlength\tabcolsep{2pt}
    \resizebox{\columnwidth}{!}{%
    \begin{tabular}{lccccccccc}
    \toprule
    \textbf{Methods} & \multicolumn{3}{c}{\textbf{AriaMultiagent}} & & \multicolumn{5}{c}{\textbf{ReplicaMultiagent}} \\
    \cmidrule(lr){2-4} \cmidrule(lr){6-10}
    & \textbf{Room0} & \textbf{Room1} & \textbf{Avg.} & & \textbf{Off0} & \textbf{Apt0} & \textbf{Apt1} & \textbf{Apt2} & \textbf{Avg.} \\
    \midrule
    ORB-SLAM3~\cite{Campos_2021} & 3.18 & 2.85 & 3.01 & & 0.60 & 1.07 & 4.94 & 1.36 & 1.99 \\
    Gaussian-SLAM~\cite{yugay2024gaussianslamphotorealisticdenseslam} & \redx & \redx & \redx & & \nd0.33 & 0.41 & 30.13 & 121.96 & 38.21 \\
    MonoGS~\cite{Matsuki:Murai:etal:CVPR2024} & \nd1.90 & \rd2.71 & \rd2.30 & & \rd0.38 & \nd0.21 & \rd3.33 & \nd0.54 & \rd1.15 \\
    \ours w.o. LC & \rd2.13 & \nd1.24 & \nd1.69 & & 0.42 & \rd0.38 & \nd0.54 & \rd0.66 & \nd0.50 \\
    \ours & \fs1.15 & \fs0.65 & \fs0.90 & & \fs0.27 & \fs0.16 & \fs0.26 & \fs0.32 & \fs0.25 \\
    \bottomrule
    \end{tabular}%
    }
    \caption{\textbf{Tracking performance compared to single-agent methods}(ATE RMSE [cm]$\downarrow$). We compare our method with traditional and 
    state-of-the-art Gaussian-based single-agent SLAM systems. Our method outperforms all the baselines even without loop closure (LC).}
    \label{tab:tracking_vs_single}
    \vspace{-15pt}
\end{table}
\subsection{Rendering Performance}
\begin{figure*}[ht!] \centering
    \newcommand{\wratio}{0.3}
    \newcommand{\imgheight}{2.8cm}
    
    \makebox[0.01\textwidth][c]{}
    \makebox[\wratio\textwidth]{\normalsize CP-SLAM\cite{zhu2022nice}}
    \makebox[\wratio\textwidth]{\normalsize \ours (ours)}
    \makebox[\wratio\textwidth]{\normalsize Ground-Truth}
    \\
    \raisebox{\height}{\makebox[0.01\textwidth]{\rotatebox{90}{\hspace{0pt}\normalsize Apt-0}}}    
    \includegraphics[width=\wratio\textwidth,height=\imgheight]{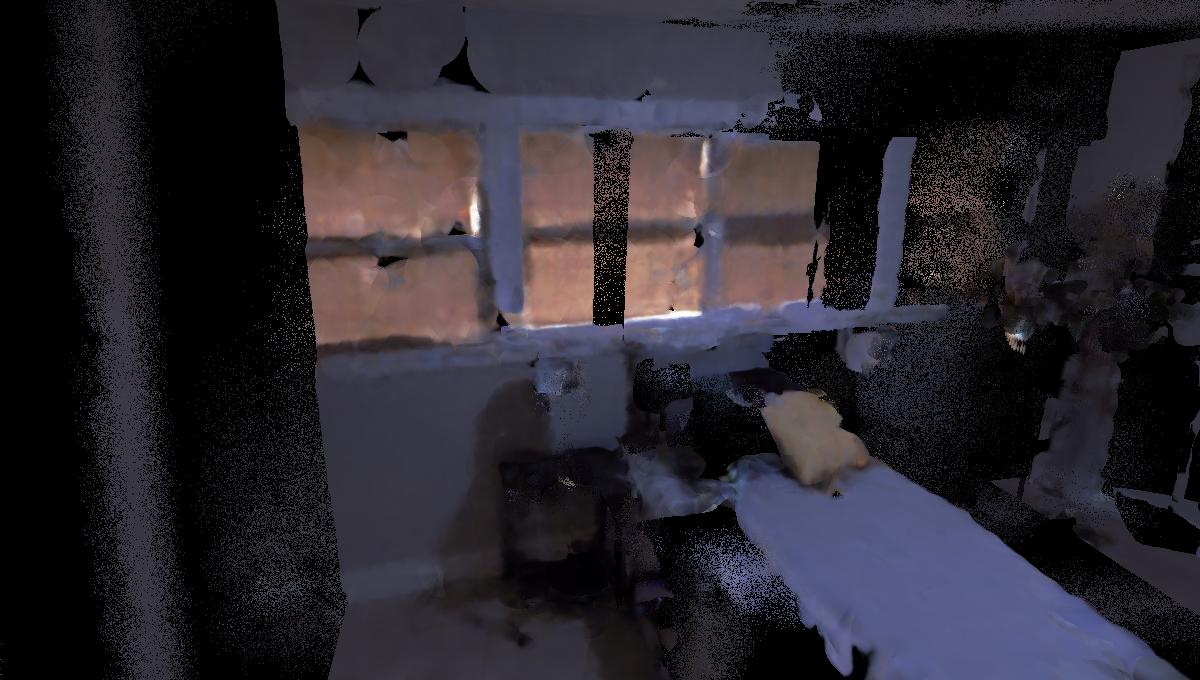}
    \includegraphics[width=\wratio\textwidth,height=\imgheight]{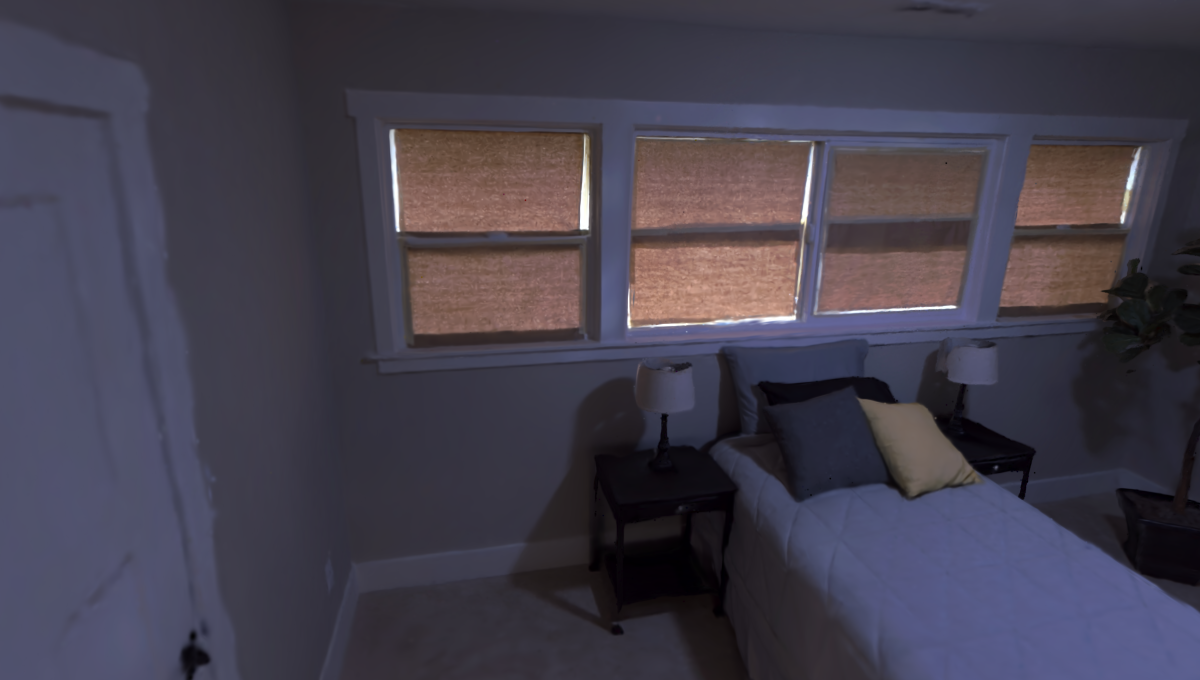}
    \includegraphics[width=\wratio\textwidth,height=\imgheight]{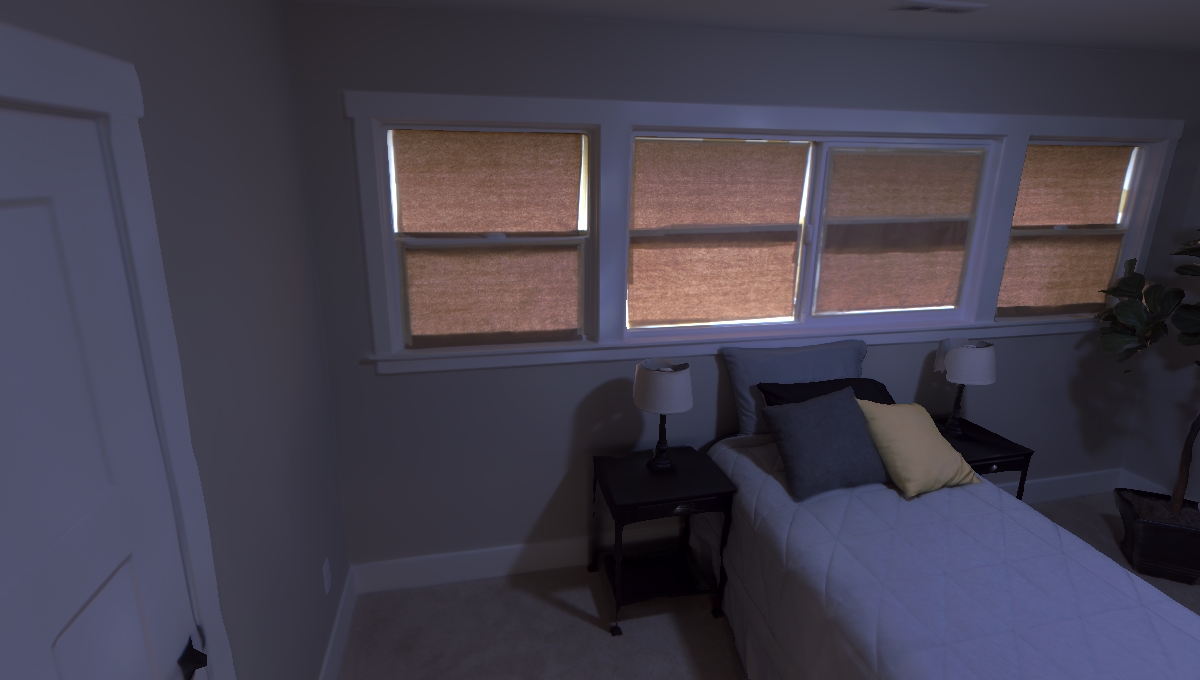}
    \\
    \raisebox{\height}{\makebox[0.01\textwidth]{\rotatebox{90}{\hspace{0pt}\normalsize Apt-0}}}    
    \includegraphics[width=\wratio\textwidth,height=\imgheight]{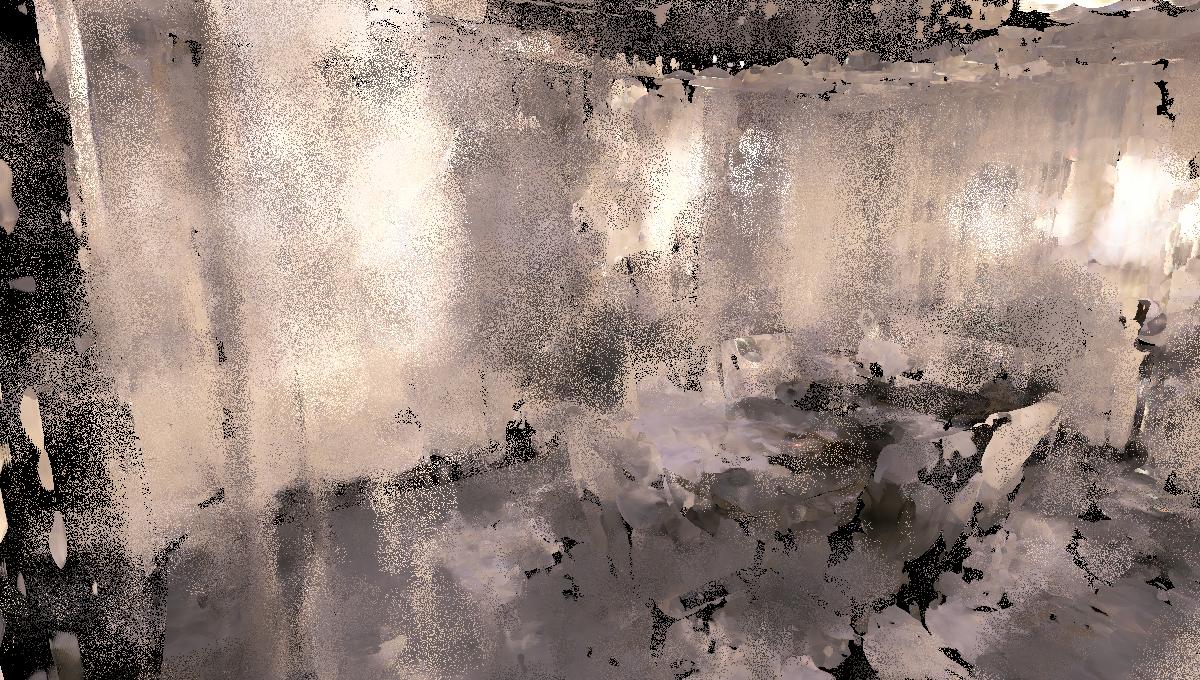}
    \includegraphics[width=\wratio\textwidth,height=\imgheight]{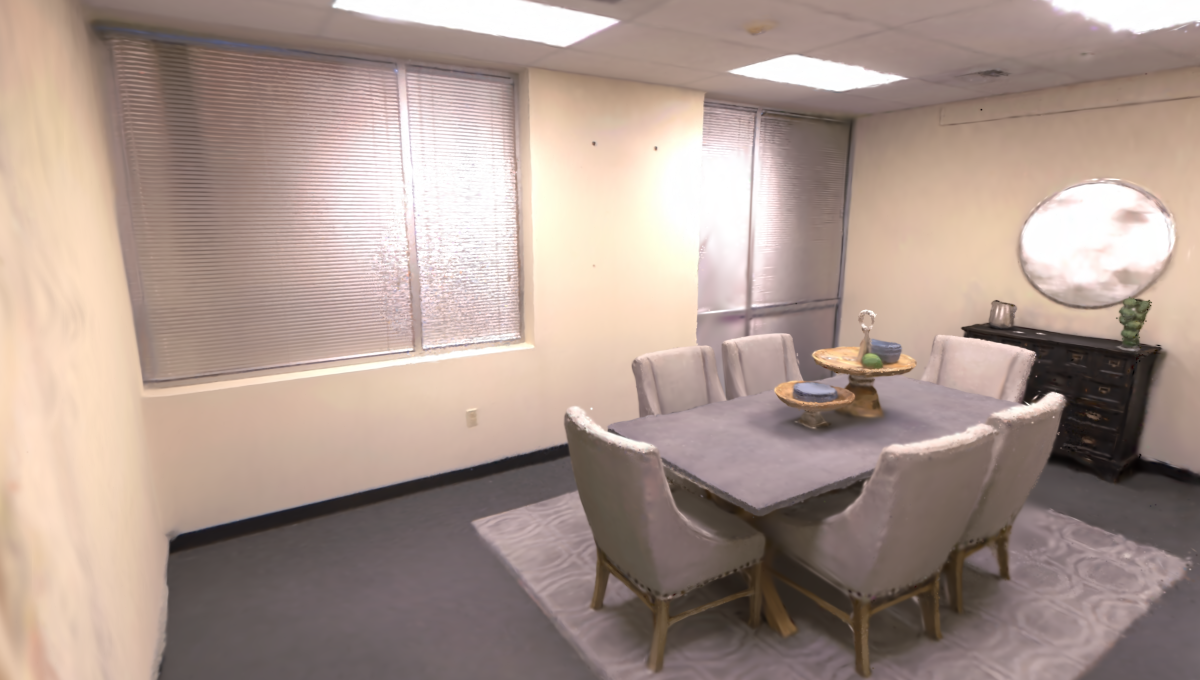}
    \includegraphics[width=\wratio\textwidth,height=\imgheight]{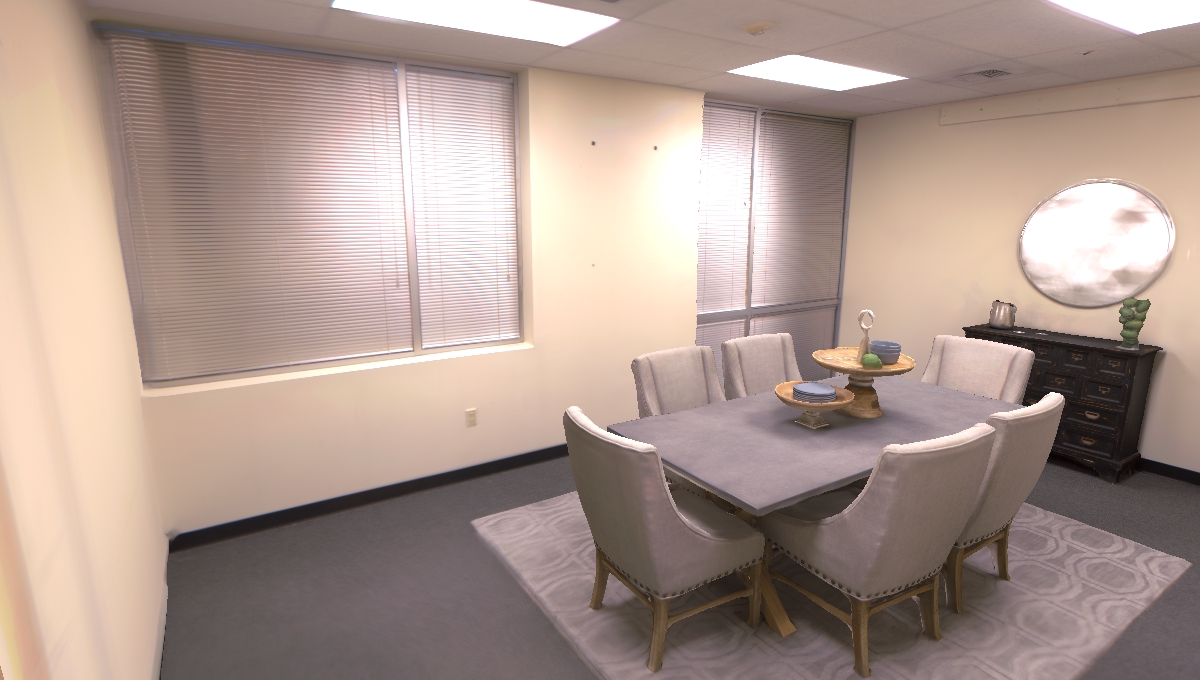}
    \\
    \raisebox{\height}{\makebox[0.01\textwidth]{\rotatebox{90}{\hspace{0pt}\normalsize Apt-2}}}    
    \includegraphics[width=\wratio\textwidth,height=\imgheight]{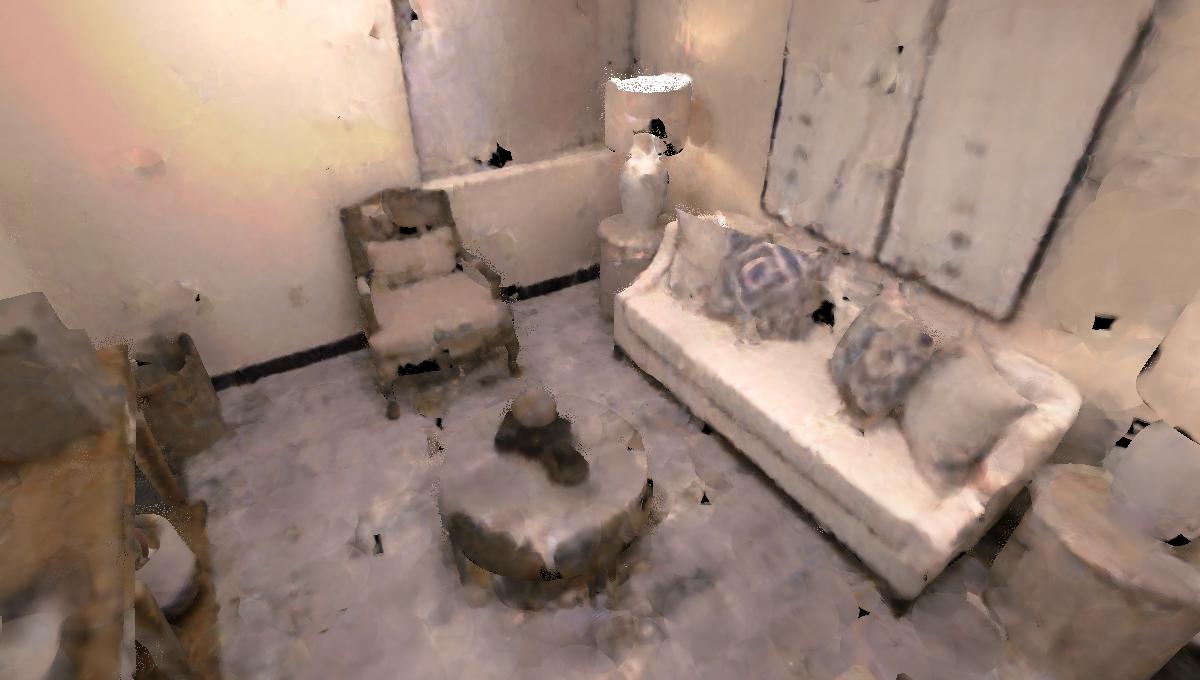}
    \includegraphics[width=\wratio\textwidth,height=\imgheight]{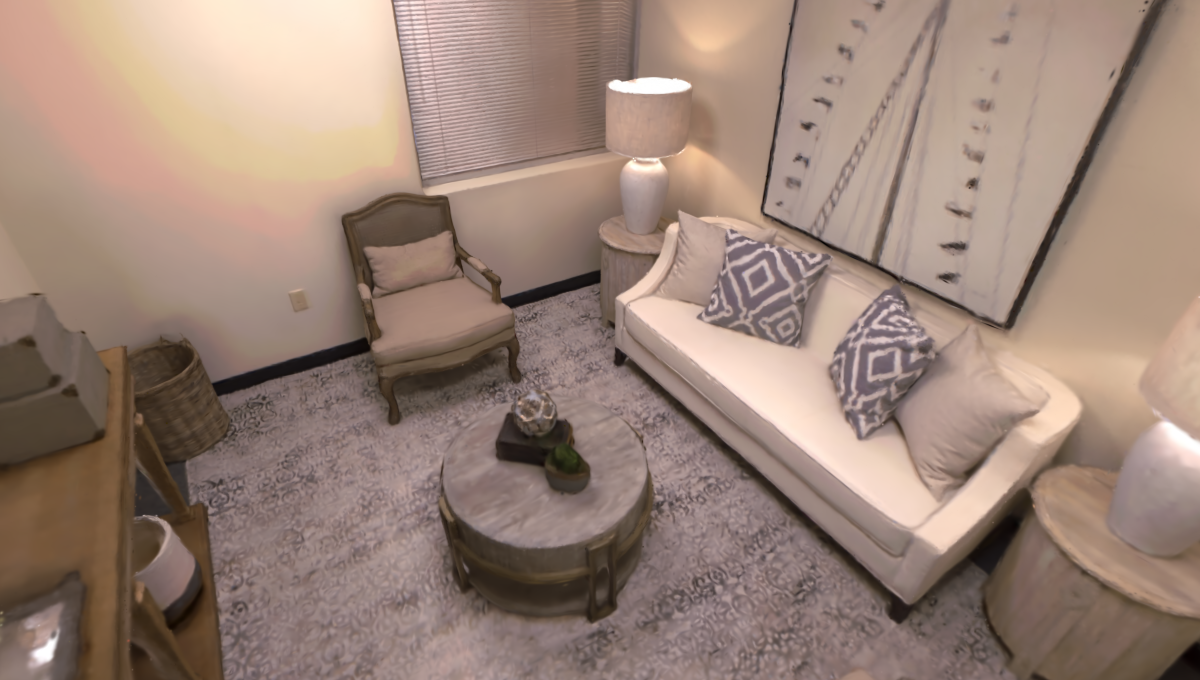}
    \includegraphics[width=\wratio\textwidth,height=\imgheight]{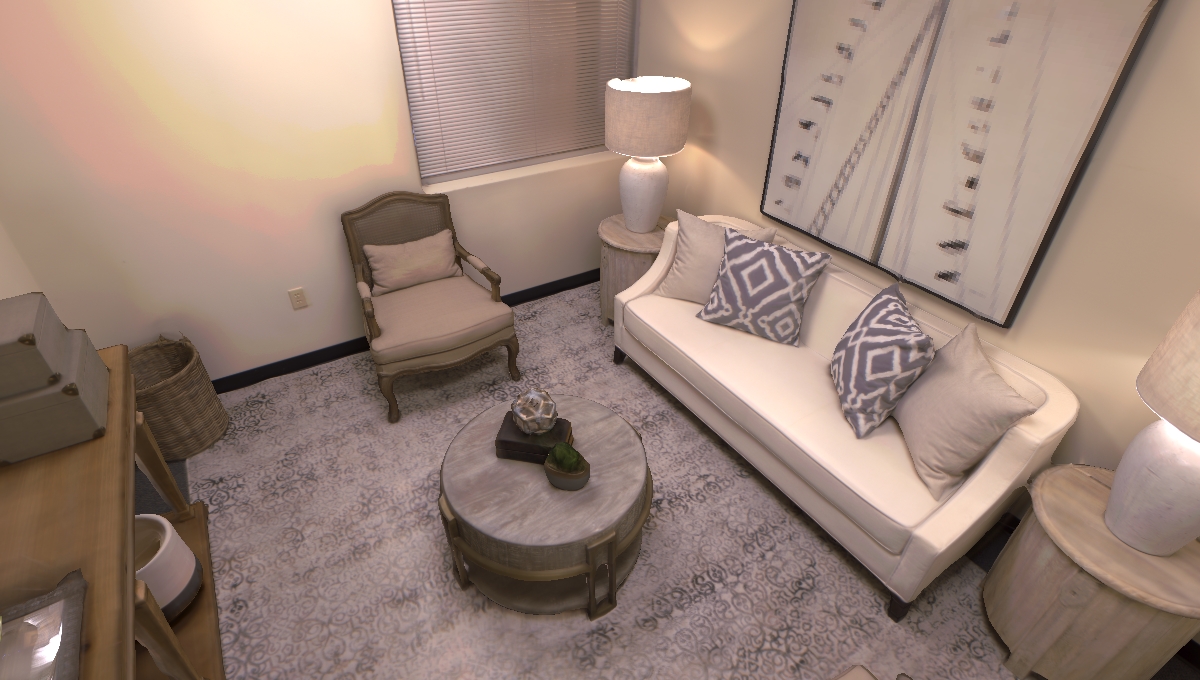}

    \caption{\textbf{Rendering performance on ReplicaMultiagent~\cite{hu2023cpslamcollaborativeneuralpointbased}}. Thanks to GS scene representation and effective merging strategy, \ours encodes more high-frequency details and substantially increases the quality of the renderings.}
    \label{fig:rendering_qualitative}
\end{figure*}

We evaluate the rendering performance on the merged scene obtained
from all the agents on ReplicaMultiagent in Table~\ref{tab:rendering_replica_multi} and AriaMultiagent datasets in Table~\ref{tab:rendering_aria_multi}.
Since CP-SLAM does not support the operation of more than two agents, we report its results on the first two out of three agents in the AriaMultiagent dataset. Thanks to 3D Gaussian splatting, agents' sub-maps can accurately render real-world data. Moreover, they are efficiently corrected using optimized poses from the loop closure module. Combined with an effective map-merging strategy, this approach allows our method to significantly outperform previous work in rendering both real-world training data and novel views. We also provide qualitative results in Fig.~\ref{fig:rendering_qualitative}.

\begin{table}[ht]
    \centering
    \setlength\tabcolsep{4pt}
    \resizebox{\columnwidth}{!}{%
    \begin{tabular}{llrrrrr}
    \toprule
    \textbf{Methods} & \textbf{Metrics} & \textbf{Off-0} & \textbf{Apt-0} & \textbf{Apt-1} & \textbf{Apt-2} & \textbf{Avg.} \\
    \midrule
    \textbf{CP-SLAM \cite{hu2023cpslamcollaborativeneuralpointbased}}
    & PSNR [dB] $\uparrow$       & 28.56 & 26.12 & 12.16 & 23.98 & 22.71 \\
    & SSIM $\uparrow$            & 0.87 & 0.79 & 0.31 & 0.81 & 0.69 \\
    & LPIPS $\downarrow$         & 0.29 & 0.41 & 0.97 & 0.39 & 0.51 \\
    & Depth L1 [cm] $\downarrow$ & 2.74 & 19.93 & 66.77 & 2.47 & 22.98 \\
    \midrule
    \textbf{\ours}
    & PSNR [dB] $\uparrow$       & 39.32 & 36.96 & 30.01 & 30.73 & \textbf{34.26} \\
    & SSIM $\uparrow$            & 0.99 & 0.98 & 0.95 & 0.96 & \textbf{0.97} \\
    & LPIPS $\downarrow$         & 0.05 & 0.09 & 0.18 & 0.17 & \textbf{0.12} \\
    & Depth L1 [cm] $\downarrow$ & 0.41 & 0.64 & 3.16 & 0.99 & \textbf{1.30} \\
    \bottomrule
    \end{tabular}%
    }
    \caption{\textbf{Training view synthesis performance on ReplicaMultiagent dataset.} The global map built by
    merging the maps from two agents is evaluated by synthesizing training views. Our method significantly outperforms
    previous state of the art.}
    \label{tab:rendering_replica_multi}
\end{table}

\begin{table}[ht]
    \centering
    \setlength\tabcolsep{3pt}
    \resizebox{\columnwidth}{!}{%
    \begin{tabular}{llrrrcrrr}
    \toprule
    \textbf{Methods} & \textbf{Metrics} & \multicolumn{3}{c}{\textbf{Novel Views}} & & \multicolumn{3}{c}{\textbf{Training Views}} \\
    \cmidrule(lr){3-5} \cmidrule(lr){7-9}
    & & \textbf{S1} & \textbf{S2} & \textbf{Avg.} & & \textbf{S1} & \textbf{S2} & \textbf{Avg.} \\
    \midrule
    \textbf{CP-SLAM \cite{hu2023cpslamcollaborativeneuralpointbased}}
    & PSNR [dB] $\uparrow$       & 8.96 & 9.17 & 9.06 & & 10.01 & 10.45 & 10.23 \\
    & SSIM $\uparrow$            & 0.32 & 0.24 & 0.28 & & 0.24 & 0.30 & 0.27 \\
    & LPIPS $\downarrow$         & 0.91 & 0.95 & 0.93 & & 0.93 & 0.98 & 0.95 \\
    & Depth L1 [cm] $\downarrow$ & 17.14 & 13.23 & 15.18 & & 18.42 & 12.01 & 15.12 \\
    \midrule
    \textbf{\ours} & PSNR [dB] $\uparrow$ & 23.45 & 21.78 & \textbf{22.61} & & 24.11 & 26.17 & \textbf{25.14} \\
    & SSIM $\uparrow$                     & 0.89 & 0.85 & \textbf{0.87} & & 0.91 & 0.93 & \textbf{0.92} \\
    & LPIPS $\downarrow$                  & 0.22 & 0.21 & \textbf{0.22} & & 0.20 & 0.14 & \textbf{0.17} \\
    & Depth L1 [cm] $\downarrow$          & 1.33 & 4.96 & \textbf{3.15} & & 1.87 & 1.30 & \textbf{1.59} \\
    \bottomrule
    \end{tabular}%
    }
    \caption{\textbf{Novel and training view synthesis performance on AriaMultiagent dataset.} The global map built by
    merging the maps from two agents is evaluated by synthesizing novel and training views.
    Previous state-of-the-art methods struggle to render real-world data.
    \textit{Note.} CP-SLAM is evaluated only on two agents since it does not support more than two.}
    \label{tab:rendering_aria_multi}
    \vspace{-15pt}
\end{table}

\subsection{Ablation Studies}

\boldparagraph{Effect of Pose Initialization.}
In Table~\ref{tab:ablation_pose_init} we numerically evaluate our two-stage tracking mechanism on the AriaMultiagent and
ReplicaMultiagent~\cite{hu2023cpslamcollaborativeneuralpointbased} datasets.
This result highlights the importance of pose initialization for tracking accuracy.
In addition, we show that pose initialization does not add significant computational overhead.

\begin{table}[ht]
    \centering
    \setlength\tabcolsep{4pt}
    \resizebox{\columnwidth}{!}{%
    \begin{tabular}{llrr}
    \toprule
    \textbf{Methods} & \textbf{Dataset} & \textbf{\makecell[r]{Tracking \\ Frame [s]} $\downarrow$} & \textbf{\makecell[r]{Avg. ATE RMSE \\{}[cm] $\downarrow$} } \\
    \midrule
    \multirow{2}{*}{\makecell[l]{\ours \\ (w.o. Pose Initialization)}} & AriaMultiagent & \textbf{0.66} & 0.874 \\
     & ReplicaMultiagent & \textbf{0.68} & 0.825 \\
    \midrule
    \multirow{2}{*}{\ours} & AriaMultiagent & 0.67 & \textbf{0.670} \\
     & ReplicaMultiagent & 0.69 &  \textbf{0.365} \\
    \bottomrule
    \end{tabular}%
    }
    \caption{\textbf{Ablation on pose initialization on AriaMultiagent and ReplicaMultiagent datasets}.
    Our method benefits from robust pose initialization without much computational overhead.}
    \label{tab:ablation_pose_init}
    \vspace{-10pt}
\end{table}

\boldparagraph{Loop Closure Detection.}
We evaluate the performance of our loop closure detection against the NetVLAD-based loop closure detection mechanism
proposed in~\cite{hu2023cpslamcollaborativeneuralpointbased}. For this, we integrate a NetVLAD-based loop closure detection module into our pipeline keeping all the thresholds from~\cite{hu2023cpslamcollaborativeneuralpointbased}. Thanks to the huge amount of data DinoV2~\cite{oquab2024dinov2learningrobustvisual} was trained, its features can accurately encode a large variety of data. This leads to more accurate loop closure detection as shown in Table~\ref{tab:ablation_loop_detection}.

\begin{table}[ht]
    \centering
    \resizebox{\columnwidth}{!}{%
    \begin{tabular}{llc}
    \toprule
    \textbf{Methods} & \textbf{Datasets} & \textbf{Avg.} \\
    \midrule
    \ours~(w. NetVLAD~\cite{netvlad}) & AriaMultiagent & 1.363 \\
    \ours~(w. NetVLAD~\cite{netvlad}) & ReplicaMultiagent & 0.351 \\
    \midrule
    \ours & AriaMultiagent & \textbf{0.900} \\
    \ours & ReplicaMultiagent & \textbf{0.252} \\
    \bottomrule
    \end{tabular}%
    }
    \caption{\textbf{Ablation on loop detection mechanism on ReplicaMultiagent and AriaMultiagent datasets} (ATE RMSE [cm]$\downarrow$).
    Our foundation vision model-based loop detection mechanism generalizes better than Net-VLAD descriptors to unseen data.}
    \label{tab:ablation_loop_detection}
\end{table}

\boldparagraph{Memory and Runtime Analysis.} In Table ~\ref{tab:runtime}, we compare runtime and memory usage against the most recent multi-agent neural SLAM
system~\cite{hu2023cpslamcollaborativeneuralpointbased}. Specifically, we evaluate the time required for tracking, mapping, map merging,
and peak VRAM consumption. Using Gaussian sub-maps for agent mapping and tracking significantly accelerates the pipeline and limits the VRAM required per agent. Additionally, our sub-map caching and merging strategies reduce the merging time.

\begin{table}[t]
    \centering
    \setlength\tabcolsep{3pt}
    \resizebox{\columnwidth}{!}{%
    \begin{small} %
    \begin{tabular}{lrrrr}
    \toprule
    \textbf{Methods} & \makecell{\textbf{Mapping} \\ \textbf{Frame [s]}} & \makecell{\textbf{Tracking} \\ \textbf{Frame [s]}} & \makecell[r]{\textbf{Map Merging} \\ \textbf{[s]}} & \makecell[r]{\textbf{Peak GPU} \\ \textbf{[GiB]}} \\
    \midrule
    CP-SLAM \cite{hu2023cpslamcollaborativeneuralpointbased} & 16.95 & 3.36 & 1448 & 7.70 \\
    \ours & \textbf{0.71} & \textbf{0.69} & \textbf{167} & \textbf{1.12} \\
    \bottomrule
    \end{tabular}
    \end{small} %
    }
    \caption{\textbf{Runtime and Memory Analysis on ReplicaMultiagent} \texttt{office0}. Mapping, tracking, and peak GPU consumption are computed per agent. Map merging is the time
    required to construct the global map from the agents' sub-maps.
    All metrics are computed using an NVIDIA RTX A6000 GPU.}
    \label{tab:runtime}
    \vspace{-15pt}
\end{table}
\boldparagraph{Limitations.} As shown in Table~\ref{tab:runtime}, \ours is significantly faster than the previous state-of-the-art in tracking, mapping, and global map reconstruction. However, the system operates slightly faster than 1 FPS. The implicit tracking mechanism is accurate but requires many iterations to converge. Future work could investigate solutions that enable faster convergence without compromising the accuracy of the current pipeline.

\section{Conclusion}
We present \ours, a multi-agent SLAM with novel view synthesis capabilities. Thanks to our tracking and loop closure modules, \ours demonstrates superior tracking accuracy on both synthetic and real-world datasets. Our efficient map-merging strategy allows high-quality rendering in various scenarios. The Gaussian-based scene representation significantly reduces processing time, disk, and VRAM consumption compared to the previous state of the art. Finally, our method can handle a varied number of simultaneously operating agents.

\boldparagraph{Acknowledgements.} This work was supported by TomTom, the University of Amsterdam, and the allowance of Top Consortia for Knowledge and Innovation (TKIs) from the Netherlands Ministry of Economic Affairs and Climate Policy.

{
    \small
    \bibliographystyle{ieeenat_fullname}
    \bibliography{main}
}

\clearpage

\setcounter{page}{1}
\setcounter{section}{0}
\counterwithin{figure}{section} %
\counterwithin{table}{section}
\renewcommand{\thesection}{\Alph{section}} %
\maketitlesupplementary

\begin{abstract}
    This supplementary material includes a video of \ours running on multiple agents in a real-world indoor environment, showcasing the effectiveness of tracking, mapping, loop closure, and map merging modules. Furthermore, we provide implementation details, ablations on efficiency, pose initialization, and loop constraints estimation.
\end{abstract}

\section{Video}
We provide the video $\textit{magic\_slam.mp4}$ in the supplementary material. In the video, we overview the high-level architecture of our method. We also showcase \ours online tracking and reconstruction capabilities on the AriaMultiagent \textit{room0} sequence. The video highlights the effectiveness of our globally consistent reconstruction process. It visualizes how the agents explore the environment and their estimated trajectories. For clarity, the cached sub-maps are visualized as meshes. Additionally, the video demonstrates the loop edges connecting nodes in the agents' trajectories and shows how sub-maps are updated using the optimized poses from pose graph optimization. Finally, it presents the optimized trajectories and the resulting merged global map.

\section{Implementation Details}

\boldparagraph{Hyperparameters.} \cref{tab:supp_configs} lists the hyperparameters used in our system, including $\lambda_c$ in the tracking loss, learning rates $l_r$ for rotation and $l_t$ for translation, and the number of optimization iterations $\text{iter}_t$ for tracking and $\text{iter}_m$ for mapping on the reported ReplicaMultiagent~\cite{hu2023cpslamcollaborativeneuralpointbased} and AriaMultiagent datasets. Additionally we set $\lambda_\text{color}$, $\lambda_\text{depth}$, and $\lambda_\text{reg}$ to 1 in the mapping loss $\mathcal{L}_\text{render}$ for all datasets.

\begin{table}[ht]
\centering
\small %
\setlength\tabcolsep{14pt}
\begin{tabular}{lrr}
\toprule
\textbf{Params} & ReplicaMultiagent & AriaMultiagent  \\
\midrule
$\lambda_c$      & 0.95     & 0.6    \\
$l_r$            & 0.0002   & 0.002  \\
$l_t$            & 0.002    & 0.01   \\
$\text{iter}_t$  & 60       & 200    \\
$\text{iter}_m$  & 100      & 100    \\
\bottomrule
\end{tabular}
\caption{\textbf{Per-dataset Hyperparameters} for tracking and mapping modules.}
\label{tab:supp_configs}
\end{table}

\boldparagraph{Tracking.} The inlier mask $M_\text{inlier}$ in the tracking loss filters pixels with depth errors $50$ times larger than the median depth error of the current re-rendered depth map. Pixels without valid depth input are also excluded as the inconsistent re-rendering in those areas can hinder the pose optimization. For the soft alpha mask, we adopt $M_\text{alpha} = \alpha^3$ for per-pixel loss weighting, where $\alpha$ refers to the Gaussian opacity value.

\boldparagraph{Mapping.} A new submap is triggered every 50 frames for ReplicaMultiagent and 20 frames for AriaMultiagent. We do this to synchronize the communication between the agents and the server. When selecting candidates to add to the submap at a new keyframe, we uniformly sample $M_k$ points from pixels that meet either the alpha value condition or the depth discrepancy condition. $M_k$ is set to $60K$ for ReplicaMultiagent and $100K$ for AriaMultiagent. The threshold $\alpha_\text{thre}$ is set to 0.98 across all datasets. The depth discrepancy condition masks pixels where the depth error exceeds 40 times the median depth error of the current frame. Newly added Gaussians are initialized with opacity values 0.5 and their initial scales are set to the nearest neighbor distances within the submap. After the mapping optimization for the new keyframe, we prune Gaussians with opacity values lower than a threshold $o_\text{thre}$. We set $o_\text{thre}=0.1$ for all the datasets. Upon completing the mapping and tracking of all frames for the agents' input sequences, we merge the saved sub-maps into a global map. The mesh is extracted via TSDF fusion~\cite{curless1996volumetric} using the rendered depth maps and estimated poses from the sub-maps. We perform color and depth refinement on the global map for $3K$ iterations using the same losses and hyperparameters as in the mapping stage.

\boldparagraph{Loop Closure.} We use DinoV2~\cite{oquab2024dinov2learningrobustvisual} based on the small ViT architecture for loop closure. We set feature threshold $\theta_{\text{feature}}$ to 0.35 for both ReplicaMultiagent and AriaMultiagent datasets. In the pose graph, we set the information matrix $\Omega_{ij}$ to identity for both odometry and loop closure edges. We perform pose graph optimization at the end of the run using the \textit{g2o}~\cite{g2o} library.

\section{Efficiency Evaluation}
We numerically evaluate the efficiency of our novel mapping mechanism, comparing it with Gaussian-SLAM~\cite{yugay2024gaussianslamphotorealisticdenseslam}, which was the first to incorporate sub-maps in a 3DGS SLAM pipeline. Unlike Gaussian-SLAM, our approach reduces the sub-map size on disk by avoiding caching all Gaussians. Additionally, we initialize sub-maps with significantly fewer seeded Gaussians. Gaussian-SLAM does not support multi-agent operations, so we evaluate its performance using single-agent sequences from the ReplicaMultiagent~\cite{hu2023cpslamcollaborativeneuralpointbased} dataset.

\begin{table}[ht]
    \centering
    \small %
    \setlength\tabcolsep{4pt}
    \begin{tabular}{lrr}
    \toprule
    \textbf{Methods} & \textbf{Peak disk [MB]$\downarrow$} & \textbf{Peak GPU} \textbf{[GiB]$\downarrow$} \\
    \midrule
    Gaussian-SLAM~\cite{yugay2024gaussianslamphotorealisticdenseslam} & 181 & 4.16 \\
    \ours & \textbf{54} & \textbf{1.12} \\
    \bottomrule
    \end{tabular}%
    \caption{\textbf{Sub-map disk and GPU memory ablation on ReplicaMultiagent}. Thanks to our novel mapping mechanism,
        \ours consumes more than three times less VRAM and disk space to process and store a single sub-map. All metrics are profiled using an NVIDIA H100 GPU.}
\label{tab:sup:submap_memory}
\end{table}

\section{Pose Initialization}
When selecting a tracking mechanism for our method, we experimented with both implicit and explicit approaches. In the paper, we have shown that implicit tracking with initialization used in our method is more accurate than explicit tracking. Moreover, we found that our pose initialization mechanism is not always beneficial for explicit tracking. For this experiment, we integrated our pose initialization module into the explicit tracking pipeline. We chose an explicit tracking approach from MonoGS~\cite{Matsuki:Murai:etal:CVPR2024} since it was the most accurate 3DGS-SLAM pipeline by the time of the submission~\cite{tosi2024nerfs3dgaussiansplatting}. MonoGS does not support multiple agents, therefore we ran our experiments on a single agent from AriaMultiagent dataset.

\begin{table}[ht]
    \centering
    \small %
    \setlength\tabcolsep{22pt}
    \begin{tabular}{@{}lrr@{}}
    \toprule
    \textbf{Method} & \textbf{Room-0} & \textbf{Room-1} \\
    \midrule
    Explicit tracking w.o. PI  & \textbf{0.85} & \textbf{1.65} \\
    Explicit tracking w. PI & 1.79 & 3.21 \\
    \bottomrule
    \end{tabular}%
    \caption{\textbf{Pose initialization ablation on AriaMultiagent dataset}
    (ATE RMSE [cm]$\downarrow$). We show that our pose initialization (PI) strategy is not always beneficial for explicit tracking. We integrate our pose initialization module to the existing pipeline~\cite{Matsuki:Murai:etal:CVPR2024} and evaluate its performance on single agent sequences from the AriaMultiagent dataset.}
    \label{tab:sup:pose_init}
\end{table}

\section{Loop Detection Feature Extractors}
We evaluated the inference speed of the ViT-small DinoV2~\cite{oquab2024dinov2learningrobustvisual} feature extractor, used in our pipeline, compared to the NetVLAD~\cite{netvlad} feature extractor. The same NetVLAD checkpoint was used in ~\cite{hu2023cpslamcollaborativeneuralpointbased}. DinoV2 demonstrates better generalization due to its training data without compromising speed.

\begin{table}[ht]
\centering
\small %
\setlength\tabcolsep{19pt}
\begin{tabular}{lr}
\toprule
\textbf{Method} & \textbf{Inference Time [s]$\downarrow$}  \\
\midrule
NetVlad~\cite{netvlad}      & 0.045    \\
DinoV2 ViT-small~\cite{oquab2024dinov2learningrobustvisual}            & \textbf{0.028}   \\
\bottomrule
\end{tabular}

\caption{\textbf{Inference Speed Comparison for Loop Detection Feature Extractors.} We compare the inference time of ViT-small DinoV2~\cite{oquab2024dinov2learningrobustvisual} and NetVLAD~\cite{netvlad} feature extractors, highlighting DinoV2's improved performance without compromising on speed. We use the same NetVLAD checkpoint as in ~\cite{hu2023cpslamcollaborativeneuralpointbased}. The reported numbers are averaged over the ReplicaMultiagent office0. All evaluations are done on RTX3090.}
\label{tab:supp_extractors}
\end{table}

\section{Loop Constraints Estimation}
We compute loop constraints by registering full-resolution point clouds from the first frames of the respective sub-maps. This method outperforms registering 3D Gaussian means, as the distributions of 3D Gaussian means in loop sub-maps often vary significantly between agents and are more sparse. These variations lead to less accurate loop constraints, ultimately reducing the performance of pose graph optimization.

\begin{table}[ht]
    \centering
    \small %
    \setlength\tabcolsep{3pt}
    \begin{tabular}{lr}
    \toprule
    \textbf{Method} & \textbf{ATE [cm]$\downarrow$} \\
    \midrule
    \ours with Gaussian Means Registration & 5.621 \\
    \ours with Point Cloud Registration & \textbf{0.985}\\
    \bottomrule
    \end{tabular}%
    \caption{\textbf{Loop Constraints Ablation on AriaMultiagent} \textit{room0}. We compare the impact of registering the means of sub-map 3D Gaussians across multiple agents with registering input point clouds from the first frame of each sub-map. Thanks to the similarly higher resolution of the input point clouds, point cloud registration is more accurate.}
    \label{tab:sup:pose_init}
\end{table}

\end{document}